\documentclass{IEEEtran}
\usepackage{cite}

\ifCLASSINFOpdf
  \usepackage[pdftex]{graphicx}
\else
\fi

\usepackage{amsmath}
\newtheorem{theorem}{Theorem}
\usepackage{mathtools}
\usepackage{algorithmic}
\usepackage{algorithm}
\begin{document}

\title{True Online TD-Replan($\lambda$)\\ Achieving Planning through Replaying}


\author{Abdulrahman~Altahhan
\thanks{ }
}

\markboth{IEEE International Joint Conference on Neural Networks}%
{Altahhan \MakeLowercase{\textit{et al.}}:  TD-Replan(λ) Learning: An Efficient and Scalable Replay Method}

\maketitle

\begin{abstract}
In this paper, we develop a new planning method that extends the capabilities of the true online TD to allow an agent to efficiently replay all or part of its past experience, online in the sequence that they appear with, either in each step or sparsely according to the usual $\lambda$ parameter. In this new method that we call True Online TD-Replan($\lambda$), the $\lambda$ parameter plays a new role in specifying the density of the replay process in addition to the usual role of specifying the depth of the target’s updates. We demonstrate that, for problems that benefit from experience replay, our new method outperforms true online TD($\lambda$), albeit quadratic in complexity due to its replay capabilities. In addition, we demonstrate that our method outperforms other methods with similar quadratic complexity such as Dyna Planning and TD(0)-Replan algorithms. We test our method on two benchmarking environments, a random walk problem that uses simple binary features and on a myoelectric control domain that uses both simple sEMG features and deeply extracted features to showcase its capabilities. 

\end{abstract}


%
\IEEEpeerreviewmaketitle

\section{Introduction}
%
%
%
%
\IEEEPARstart{E}{xperience} replay plays an important role in the context of reinforcement learning algorithms.
In this paper we tackle the issue of building a robust method that allows the agent to maximize its experience replay capability with relatively cheap complexity. We will tackle multi-step sequential replay algorithm where the agent replays a sequence of past experience steps in the order they appeared with. This issue have been partially attempted in \cite{TD-Replan(0)} where the algorithm used TD(0) update rules as its basis. In this work, we will extend the ideas developed in \cite{TD-Replan(0)} to the true online TD($\lambda$) updates. In particular, we will build a new method based on three requirements. First, we would like to be able to utilise a multi-step targets for each replay update instead of the one step target update of TD(0), this allows the method to choose how deep its targets are going to be for each replay update. The second requirement, is that we want to allow the algorithm designer to choose how much of the past experience updates he/she needs to incorporate and how frequently this replay process will be performed. Thirdly, regardless of the replay depth and the target’s depths, the method should be efficient even on the limit when the replay depth is maximal. 
To achieve these goals we first introduce a method, namely online $\lambda$-return TD-Replan, that takes experience replay to its extreme by allowing the agent to replay all of its past experience online in every time step. Unlike previous work, this method allows us to utilize the multi-step interim $\lambda$-return targets for each replay update instead of the one-step target of  TD($0$). We show how to deduce an online efficient incremental method, namely TD($\lambda$)-Replan, that is equivalent to online $\lambda$-return TD-Replan, but has a complexity that is not related to the time step, we prove the equivalency mathematically in Theorem 1. We show then that the true online TD algorithm becomes a special case of the true online TD-Replan algorithm in Theorem 2. We then soften the replay maximal replay requirement by utilizing a parameterization that interpolates between no-replay and replay-all, effectively allowing the designer to choose the online replay depth similar to how we choose the depth of $\lambda$-return in the true online TD($\lambda$). Finally, the target's depth and the replay process depth of the resultant TD($\lambda$)-Replan($\acute{\lambda}$) method, can be aligned to formulate TD-Replan($\lambda$).

Reinforcement learning deals with Markov Decision Process $(\mathcal{S,A},p,r,\gamma)$ where we have a set of discrete actions $\mathcal{A}$ that are available for the agent to pick from at each state $s$. We denote the feature representation of state $s$ as ${\boldsymbol{\phi}}(s)$ and the number of features $|{\boldsymbol{\phi}}|=n$. $r(s,a,\acute{s})$ is the expected reward signal for taking action $a$ at state $s$ then moving to state $\acute{s}$. The feature representation can be complex. For example, the features can be obtained from a deep architecture such as from the encoder of an auto-encoder \cite{LearningDeepArch} or a set of stacked auto encoders \cite{OwnStacked} . We can then employ whatever algorithm we have on the resultant features \cite {SuttonBarto}. In this case each learning stage is dissected from the other stages. In this paper we take this approach due to its simplicity and theoretical guarantees. Alternatively, one can take a more integrated end-to-end learning approach. An example of such integrated approach is to build a deep feature extraction model and append a fully connected layer with linear activation function, we then can backpropagate the error coming out of the last layer further into previous layers \cite{NatureMnih}.
However, despite the impressive empirical achievement of such models convergence guarantees do not apply straight on a non-linear model \cite{SuttonBarto}. The high performance in the second approach can mainly be attributed to the stability and agility provided by the replay process \cite{Li2018DeepRL}. By building the targets in the method itself we conjecture that we can avoid having to use a separate network to represents the targets as in \cite{NatureMnih}. The better performance of an intermediate replay depth value can be attributed to the ability to balance the importance of sequence of events and to break the strong correlation that can lead to instability when non-linear function approximation is used. In this work we build an algorithm that is particularly suitable for deep learning whether the learning process is dissected or integrated.

Throughout our presentation we will assume that the behaviour policy is derived directly from the learning weights for action-values and we do not consider policy gradient methods. Our focus will be on multi-steps policy evaluation methods that involves looking back at multiple updates for steps that happened in the past and repeat their respective updates, i.e. replay them. 


\section{Bundled Experience Replay}
Replay can be categorized as sequential with specific frequency (ex. replaying past sequence of 10 steps every other step), which is the topic of this paper, and non-sequential, for example the one used in \cite{LinReplay}. When sequential replay is incorporated in the method we can systematically reduce its complexity to be related to the number of used features, whether a deep or a shallow model is used, rather than in the number of time steps of an experience. More recently, the replay process has been taken from a different perspective through the work of \cite{DeeperExpReplay} to mean re-evaluating past target updates using what has been recently learned by the agent, they tagged it as replay. 
In this sense their algorithms redo the same set of past updates, with the same initial learning weights in each step, using updated targets, in order to benefit from past experience. In their setting each bundle of updates always starts from the same initial weight’s values, which is a key issue. Although this simplifies finding incremental forms for the learning process, however in that sense their approach is more of reevaluating the target rather than actually replaying the experience. From our perspective, replaying the experience requires going through a bundle of past experience and redo the updates as if the agent went through them again but with its current set of weights \cite{TD-Replan(0)}. Therefore, our approach for experience replay is more like the original approach of Lin (combined with the notion of target revaluation) but contrary to Lin’s it is more sequential and more intensive to promote learning agility.  
From a learning perspective, each time a new online interaction takes place between the agent and the environment, the replay process should allow the model to start with a better initialization of the weights.
\section{TD-Replan with Interim $\lambda$-return}
Contrary to \cite{TD-Replan(0)} we use interim $\lambda$-return as the target for each update. Interim $\lambda$-returns takes advantage of all past experience to obtain a more accurate estimate of the targets of the TD updates. In this section we show the forward view of our elaborate replay method using interim $\lambda$-returns similar to the way true online TD($\lambda$) was constructed \cite{trueTD}). The forward true online TD() algorithm is largely kept as is with one important change. We will run through all past updates and redo them all as a bundle, using the latest model weights, i.e without reinitializing them back to their original values. We assume that in each time step $t$, the algorithm is going to go back to all past steps trajectory and replay every single update based on its latest weights. Index $t$ will be used to represent current time step, while index $k$ will be used to represent past steps, where $0\leq k \leq t$. For example, the model's weights at time step $t$ that are used to replay past step $k$ are denoted ${{\boldsymbol{\theta}}}_k^t$, while the weights that are the results of replaying past time step $k$ are denoted ${{\boldsymbol{\theta}}}_{k+1}^t$.  ${{\boldsymbol{\theta}}}_i^i$ will be abbreviated as ${\boldsymbol{\theta}}_i$; i.e. ${\boldsymbol{\theta}}_i \colon={\boldsymbol{\theta}}_i^i$, for example when we see  ${\boldsymbol{\theta}}_t$ it stands for  ${\boldsymbol{\theta}}_t^t$.
We will devote our attention in this section to the last layer of the model that is used to represents the value function $V$. Each one of the forward TD replay updates can be written as
\begin{equation} \label{eq:replayUpdate}
     {{\boldsymbol{\theta}}}_{k+1}^{t+1}= {{\boldsymbol{\theta}}}_k^{t+1}+\alpha_k \nabla_{\boldsymbol{\theta}} V \left(G_k^{\lambda|t+1}- V(s|{\boldsymbol{\theta}}_k^{t+1})\right) 
\end{equation}
where $G_k^{\lambda|t+1}$ is the interim $\lambda$-return introduced in \cite{trueTD} and is defined as: 
\begin{align}
    G_k^{\lambda|t}&=\sum_{i=1}^{t-k-1}{\lambda^{i-1} G_{k}^{\left(i\right)}}+\lambda^{t-k-1}G_{k}^{\left(t-k\right)}                       \label{eq:lambsReturn}\\
    G_{k}^{\left(i\right)}&=\left(1-\lambda\right)\sum_{j=1}^{i}{\gamma^{j-1}R_{k+j}}+\gamma^i V(S_{k+j}|{{\boldsymbol{\theta}}}_{k+j-1})         \label{eq:Return}
\end{align}
Note that when $k=t-1$ then $G_k^{\lambda|t}=G_{t}^{(1)}=R_{t+1} + \gamma V(S_{t+1}|\theta_{t})$ which is the usual one-step target of TD(0).
We assume that the last layer is linear, hence a linear model is used to express the value function, $V(s|{\boldsymbol{\theta}})={\boldsymbol{\theta}}^\top{\boldsymbol{\phi}}(s)$, where $\nabla_{\boldsymbol{\theta}} V={\boldsymbol{\phi}}(s)$.
This assumption entails some restriction but it does not prevent us from using a non-linear and complex layers that come before this last layer in order to build a full fledged deep learning model.

In this case, the set of the replay updates (\ref{eq:replayUpdate}) is written as
\begin{align}
\label{eq:Theta_k_update}
     {{\boldsymbol{\theta}}}_{k+1}^{t+1} &= {{\boldsymbol{\theta}}}_k^{t+1}+\alpha_k \boldsymbol{\phi}_k\left(G_k^{\lambda|t+1}-\left( {{\boldsymbol{\theta}}}_k^{t+1}\right)^\top \boldsymbol{\phi}_k\right)\\
      {{\boldsymbol{\theta}}}_{k+1}^{t+1} &=  \boldsymbol{A}_k {{\boldsymbol{\theta}}}_k^{t+1}+ \boldsymbol{b}_k^t \\
      \boldsymbol{A}_k &:=\left[{\boldsymbol{\mathcal{I}}}_{n\times n}-\alpha_k \boldsymbol{\phi}_k\boldsymbol{\phi}_k^\top\right]\\ \label{eq:defAk}
      \boldsymbol{b}_k^t &:=\alpha_k \boldsymbol{\phi}_k G_k^{\lambda|t+1} 
\end{align}
where $\boldsymbol{A}_k$ is a squared matrix, $ \boldsymbol{b}_k^t$ is a vector and $n$ is the number of weights used to encode the value function. The time and space complexity of the above algorithm can be made reasonable and be only related to $n$. Although each step is entailing $t$ updates with complexity $O(t\times n)$, we shall use the formalism used by \cite{IndepSpan} and \cite{trueTD} to make the complexity $O(n^2)$ assuming that $t>n$ or $t\gg n$ in most of the cases. 

In addition, it can be proven \cite{trueTD} that:
    \begin{align} 
    \nonumber
       G_k^{\lambda|t+1} \!- G_k^{\lambda|t} &= (\lambda \gamma)^{t-k}(R_{t+1} + \gamma{\boldsymbol{\theta}}_{t}^\top{\boldsymbol{\phi}}_{t+1} - {\boldsymbol{\theta}}_{t-1}^\top{\boldsymbol{\phi}}_{t})\\
       \label{eq:G_k defintion}
       G_k^{\lambda|t+1} &= G_k^{\lambda|t} + (\lambda \gamma)^{t-k}\acute{\delta_t}
     \end{align}
Let us see how the algorithm behaves along with the targets $G_k^{\lambda|t+1}$ in the below snippet.
\begin{align*}
    &{{\boldsymbol{\theta}}}_1^4=\boldsymbol{A}_0{{\boldsymbol{\theta}}}_3^3+\boldsymbol{b}_0^4: \quad  \boldsymbol{A}_0 ={\boldsymbol{\mathcal{I}}}-\alpha_0 \boldsymbol{\phi}_0\boldsymbol{\phi}_0^\top,\quad \boldsymbol{b}_0^4 =\alpha_0 \boldsymbol{\phi}_0 G_0^{\lambda|4}\\
    &{{\boldsymbol{\theta}}}_2^4=\boldsymbol{A}_1{{\boldsymbol{\theta}}}_1^4+\boldsymbol{b}_1^4: \quad  \boldsymbol{A}_1 ={\boldsymbol{\mathcal{I}}}-\alpha_1 \boldsymbol{\phi}_1\boldsymbol{\phi}_1^\top,\quad \boldsymbol{b}_1^4 =\alpha_1 \boldsymbol{\phi}_1 G_1^{\lambda|4}\\
    &{{\boldsymbol{\theta}}}_3^4=\boldsymbol{A}_2{{\boldsymbol{\theta}}}_2^4+\boldsymbol{b}_2^4:  \quad  \boldsymbol{A}_2 ={\boldsymbol{\mathcal{I}}}-\alpha_2 \boldsymbol{\phi}_2\boldsymbol{\phi}_2^\top,\quad \boldsymbol{b}_2^4 =\alpha_2 \boldsymbol{\phi}_2 G_2^{\lambda|4}\\
    &{{\boldsymbol{\theta}}}_4^4=\boldsymbol{A}_3{{\boldsymbol{\theta}}}_3^4+\boldsymbol{b}_3^4: \quad  \boldsymbol{A}_3 ={\boldsymbol{\mathcal{I}}}-\alpha_3 \boldsymbol{\phi}_3\boldsymbol{\phi}_3^\top,\quad \boldsymbol{b}_3^4 =\alpha_3 \boldsymbol{\phi}_3 G_3^{\lambda|4}
\end{align*}This is the same process performed in developing true online TD, except we will force each bundle of updates to initialise the first weight with the weight of the previous bundle of updates. By doing so we will be replaying all past updates in each time step.
It should be noted that if we fixed the initial weights of a bundle of update steps, then the algorithm turns into just a revaluation of all past rules instead of replaying past experience, this is what algorithms at \cite{DeeperExpReplay},\cite{IndepSpan} and \cite{trueTD} are performing, there is no replay process utilized in them.

Hence, based on our algorithm, the final weights ${{\boldsymbol{\theta}}}_4^4$ for time steps 4 can be calculated in terms of the initial weights ${{\boldsymbol{\theta}}}_3^3$ by backward substitutions as 
\begin{align*}
&{{\boldsymbol{\theta}}}_4 \!=\!{{\boldsymbol{\theta}}}_4^4\!=\!\boldsymbol{A}_3\boldsymbol{A}_2\boldsymbol{A}_1\boldsymbol{A}_0{{\boldsymbol{\theta}}}_3^3 \!+\!\boldsymbol{A}_3\boldsymbol{A}_2\boldsymbol{A}_1\boldsymbol{b}_0^4 \!+\!\boldsymbol{A}_3\boldsymbol{A}_2\boldsymbol{b}_1^4 \!+\!\boldsymbol{A}_3\boldsymbol{b}_2^4 \!+\!\boldsymbol{b}_3^4 
\end{align*}
Note that we we need $G_k^{\lambda|4}$ to be able to calculate  $\boldsymbol{b}_k^4$. 
It can be easily proven by induction that 
\begin{align}
    \label{theta_Basic_incrmental_form}
    &{{\boldsymbol{\theta}}}^{t+1}_{t+1} = \left(\prod_{i=t}^{0}\boldsymbol{A}_i\right){{\boldsymbol{\theta}}}_t^{t} + \sum_{k=0}^{t}\left(\prod_{i=t}^{k+1}\boldsymbol{A}_i\right)\boldsymbol{b}_k^{t+1}
\end{align}
It should be noted that this algorithm is completely different than the algorithm that we developed in \cite{TD-Replan(0)} in two ways; first the basic updates of each replay step is based on true online TD not on TD(0), i.e. our algorithm uses the truncated $\lambda$-return targets, the second difference is that the matrices $A_k$ are defined differently.
It should be noted that every bundle of update ends at ${{\boldsymbol{\theta}}}_{t+1}^{t+1}$ which can be used to summarize all past sequence of intermediate updates as has been shown in \cite{trueTD}. We call the above algorithm the \textit{true online $\lambda$-return TD-Replan} to emphasise two things; first that the algorithm is replaying all past experience using true online TD($\lambda$) updates, second to emphasise the link between replaying and planning which has been already established in \cite{DeeperExpReplay}. Clearly, this algorithm is expensive with a complexity of $O(n\times t)$ and in its current form it is impractical. In the next section we develop a more efficient way to achieve the same set of updates. This algorithm constitutes the non-incremental forward view of a more efficient and incremental algorithm that we call the \textit{true online TD($\lambda$)-Replan}.
\section{True Online TD($\lambda$)-Replan: An Incremental Online View}
To arrive to a correct efficient form for the intensive replay mechanism presented in the previous section, we need to extend the mathematical formulation developed in deriving the incremental forms of the true online TD($\lambda$) to accommodate the replay process. 

\begin{theorem} \label{Theorem:1}
    Given a set of n weights ${{\boldsymbol{\theta}}}$ that are due to the forward true online TD($\lambda$)-Replan algorithm shown earlier, we can obtain exactly ${{\boldsymbol{\theta}}}$ incrementally according to the following step updates
    \begin{align*}
        \boldsymbol{A}_t &=\left[{\boldsymbol{\mathcal{I}}}_{n\times n}-\ \alpha_t\boldsymbol{\phi}_t{\boldsymbol{\phi}_t}^\top\right]\\
        {\boldsymbol{e}}_t &= \boldsymbol{A}_t\gamma\lambda{\boldsymbol{e}}_{t-1}+\alpha_t\boldsymbol{\phi}_t\\
        \bar{\boldsymbol{e}}_t &= \boldsymbol{A}_t {\bar{\boldsymbol{e}}}_{t-1}+{\boldsymbol{e}}_{t}{\left(\delta_t + {{\boldsymbol{\theta}}}_{t}^\top{\boldsymbol{\phi}}_{t} - {{\boldsymbol{\theta}}}_{t-1}^\top{\boldsymbol{\phi}}_{t} \right)}+ \alpha_t{\boldsymbol{\phi}}_t{{\boldsymbol{\theta}}}_{t}^\top{\boldsymbol{\phi}}_{t}\\
        \bar{\boldsymbol{A}}_t &= \boldsymbol{A}_t {\bar{\boldsymbol{A}}}_{t-1}\\
        {{\boldsymbol{\theta}}}_{t+1} &= \bar{\boldsymbol{A}}_t{{\boldsymbol{\theta}}}_t+{\bar{\boldsymbol{e}}}_t
    \end{align*}

\end{theorem}
\begin{IEEEproof}
   We start from (\ref{theta_Basic_incrmental_form}), similar to \cite{trueTD} we denote	
     \begin{align} \label{eq:Atk1}
     \boldsymbol{A}_t^{k+1} &:= \boldsymbol{A}_t\boldsymbol{A}_{t-1}\ldots\boldsymbol{A}_{k+1} =\prod_{i=t}^{k+1}\boldsymbol{A}_i\\
    \label{eq:Att1}
        \boldsymbol{A}_t^{t+1} &:= {\boldsymbol{\mathcal{I}}}_{n\times n}\\
     \boldsymbol{A}_t^t &=\boldsymbol{A}_t
    \end{align}
    We define 
    \begin{align}
    \label{defAhat}
     {\bar{\boldsymbol{A}}}_t &:= \boldsymbol{A}_t^0\\
    \label{defz}
    {\boldsymbol{\bar{e}}}_t &:= \sum_{k=0}^{t}{\boldsymbol{A}_t^{k+1}{\boldsymbol{b}}_k^{t+1}} 
    \end{align}
    Update (\ref{theta_Basic_incrmental_form}) becomes  
     \begin{align}
        \label{eq:Theta_incremental_update}
        {{\boldsymbol{\theta}}}_{t+1}&={\bar{\boldsymbol{A}}}_t{{\boldsymbol{\theta}}}_t+{\boldsymbol{\bar{e}}}_t
    \end{align}
    Now we turn our attention to ${\boldsymbol{\bar{e}}}_t$ to come up with an incremental form to update it online. By plugging definition in (\ref{eq:G_k defintion}) in the definition of ${\boldsymbol{\bar{e}}}_t$ (\ref{defz}) and utilizing (\ref{eq:Att1}) we obtain that
     \begin{align}
        \nonumber
         {\boldsymbol{\bar{e}}}_t &:= \sum_{k=0}^{t}{\boldsymbol{A}_t^{k+1}\alpha_k{\boldsymbol{\phi}}_k G_k^{\lambda|t+1}}\\  \nonumber
          &=\sum_{k=0}^{t-1}\boldsymbol{A}_{t}^{k+1}\alpha_k{\boldsymbol{\phi}}_k G_{k}^{\lambda|t+1} + \alpha_t{\boldsymbol{\phi}}_t G_t^{\lambda|t+1}\\ \nonumber
        &=\sum_{k=0}^{t-1}\boldsymbol{A}_{t}^{k+1}\alpha_k{\boldsymbol{\phi}}_k \left(G_{k}^{\lambda|t} + (\gamma\lambda)^{t-k}\acute{\delta_t}\right) \\ \nonumber &\qquad \qquad  \qquad + \alpha_t{\boldsymbol{\phi}}_t \left( R_{t+1} + \gamma\boldsymbol{\theta}_{t}^\top \boldsymbol{\phi}_{t+1} \pm\boldsymbol{\theta}_{t-1}^\top \boldsymbol{\phi}_{t} \right)\\    \nonumber
        &=\boldsymbol{A}_{t}\sum_{k=0}^{t-1}\boldsymbol{A}_{t-1}^{k+1}\alpha_k{\boldsymbol{\phi}}_k G_{k}^{\lambda|t} + \acute{\delta_t}\sum_{k=0}^{t-1}\boldsymbol{A}_{t}^{k+1}\alpha_k{\boldsymbol{\phi}}_k(\gamma\lambda)^{t-k} \\ \nonumber
        &\qquad \qquad \qquad + \alpha_t{\boldsymbol{\phi}}_t \left( \acute{\delta_t} + \boldsymbol{\theta}_{t-1}^\top \boldsymbol{\phi}_{t} \right)\\  \nonumber
         &=\boldsymbol{A}_{t} {\boldsymbol{\bar{e}}}_{t-1} + \acute{\delta_t}\sum_{k=0}^{t-1}\boldsymbol{A}_{t}^{k+1}\alpha_k{\boldsymbol{\phi}}_k(\gamma\lambda)^{t-k} + \alpha_t{\boldsymbol{\phi}}_t \left( \acute{\delta_t} + \boldsymbol{\theta}_{t-1}^\top \boldsymbol{\phi}_{t} \right)\\
         \label{eq:z_bar}
         {\boldsymbol{\bar{e}}}_t  &=\boldsymbol{A}_{t} {\boldsymbol{\bar{e}}}_{t-1} + \acute{\delta_t} {\boldsymbol{e}}_{t}+ \alpha_t{\boldsymbol{\phi}}_t \left( \boldsymbol{\theta}_{t-1}^\top \boldsymbol{\phi}_{t} \right)\\
         {\boldsymbol{e}}_{t}&:=\sum_{k=0}^{t}(\gamma\lambda)^{t-k}\boldsymbol{A}_{t}^{k+1}\alpha_k{\boldsymbol{\phi}}_k
     \end{align}
     Note that $\alpha_t$ is included in the definition of ${\bar{\boldsymbol{e}}}_t$.
    In addition, ${\boldsymbol{e}}_{t}$  is defined as in \cite{trueTD} and hence we can readily utilize its derived incremental form
    \begin{align}\nonumber
        {\boldsymbol{e}}_t &:=\sum_{k=0}^{t}{\left(\gamma\lambda\right)^{t-k}\boldsymbol{A}_t^{k+1}\alpha_k{\boldsymbol{\phi}}_k\ }\\
           \label{eq:e}
           {\boldsymbol{e}}_t&={\boldsymbol{A}}_t\gamma\lambda {\boldsymbol{e}}_{t-1}+\alpha_t{\boldsymbol{\phi}}_t
    \end{align}
    Finally, from the definition of $\boldsymbol{A}_t^0$ in (\ref{eq:Atk1}) we have
    \begin{align}
    \label{eq:A_bar}
        {\bar{\boldsymbol{A}}}_t=\boldsymbol{A}_t\boldsymbol{A}_{t-1}\ldots\boldsymbol{A}_0=\boldsymbol{A}_t{\bar{\boldsymbol{A}}}_{t-1}
    \end{align} which constitutes the incremental form for ${\bar{\boldsymbol{A}}}_{t}$. 
 The initial conditions as per the definitions are set to 
    $\bar{\boldsymbol{A}}_{-1}={\boldsymbol{\mathcal{I}}}_{n\times n},
     \bar{\boldsymbol{A}}_{-1}={\boldsymbol{\mathcal{I}}}_{n\times n},
     ,{\boldsymbol{e}}_{-1}=\mathbf{0}_{n\times1}$ 
     which yields TD(0) update for $t=0$. Hence our algorithm is defined by (\ref{eq:defAk}), (\ref{eq:Theta_incremental_update}), (\ref{eq:z_bar}), (\ref{eq:e}) and (\ref{eq:A_bar}) which conclude our proof.
\end{IEEEproof}
\section{Efficient TD-Replan Algorithm}
By substituting $A_t$ and reorganising the terms so that we have vector $\times$ matrix multiplication and not matrix $\times$ matrix multiplication we obtain the true online TD($\lambda$)-Replan method shown below:
   \begin{align}
   \label{eq:e_t incrmental}
        {\boldsymbol{e}}_t &= \gamma\lambda{\boldsymbol{e}}_{t-1} + \alpha_t\boldsymbol{\phi}_t\left(1 - \boldsymbol{e}_{t-1}^\top\boldsymbol{\phi}_t\right)\\ \nonumber
        \label{eq:ebar_t incremental}
        \bar{\boldsymbol{e}}_t &= {\bar{\boldsymbol{e}}}_{t-1} + {\boldsymbol{e}}_{t}{\left(\delta_t + {{\boldsymbol{\theta}}}_{t}^\top{\boldsymbol{\phi}}_{t} - {{\boldsymbol{\theta}}}_{t-1}^\top{\boldsymbol{\phi}}_{t} \right)} 
        \\      &\qquad \qquad \qquad \qquad 
        - \alpha_t\boldsymbol{\phi}_t \left(  {\bar{\boldsymbol{e}}}_{t-1}^\top\boldsymbol{\phi}_t - {{\boldsymbol{\theta}}}_{t-1}^\top{\boldsymbol{\phi}}_{t} \right)\\
        \label{eq:Abar_t incremental}
        \bar{\boldsymbol{A}}_t &= {\bar{\boldsymbol{A}}}_{t-1} - \alpha_t\boldsymbol{\phi}_t \left(\boldsymbol{\phi}_t^\top{\bar{\boldsymbol{A}}}_{t-1}\right)\\
        \label{eq:theta_incrmental}
        {{\boldsymbol{\theta}}}_{t+1} &= \bar{\boldsymbol{A}}_t{{\boldsymbol{\theta}}}_t+{\bar{\boldsymbol{e}}}_t
    \end{align}
Formulating this method as a learning episodic algorithm for prediction is given in Algorithm \ref{algo:TDReplay(1)}. 

\begin{algorithm}[H]
     \caption{TD($\lambda$)-Replan(1) Learning} \label{algo:TDReplay(1)}
     \begin{algorithmic}
         \renewcommand{\algorithmicrequire}{\textbf{Input:}}
         \renewcommand{\algorithmicensure}{\textbf{Output:}}
         \REQUIRE $\alpha,\gamma,\lambda, {\boldsymbol{\theta}}_{init}$
         \ENSURE  ${\boldsymbol{\theta}}$
          \STATE obtain initial $\boldsymbol{\phi}$
          \STATE ${\boldsymbol{\theta}} \leftarrow {\boldsymbol{\theta}_{init}}$
          \FOR {all episodes}
              \STATE ${\boldsymbol{e}} \leftarrow \boldsymbol{0}, \bar{\boldsymbol{e}} \leftarrow 0, \bar{\boldsymbol{A}}  \leftarrow \mathcal{I}, V_{old}\leftarrow 0$
              \WHILE{$S$ is not Terminal}
                  \STATE obtain next feature vector  $\boldsymbol{\phi}'$ and reward $R$
                  \STATE $ V \leftarrow {\boldsymbol{\theta}}^\top\boldsymbol{\phi}$
                  \STATE $ {V'} \leftarrow {\boldsymbol{\theta}}^\top\boldsymbol{{\phi}}'$
                  \STATE $ \delta \leftarrow R + \gamma {V'} -V$
                  \STATE $
                      {\boldsymbol{e}} \leftarrow {\boldsymbol{e}}\gamma\lambda -\alpha\boldsymbol{\phi}(\gamma\lambda{\boldsymbol{e}}^\top\boldsymbol{\phi} -1)$ 
                  \STATE $
                        {\bar{\boldsymbol{e}}} \leftarrow {\bar{\boldsymbol{e}}} \quad -\alpha\boldsymbol{\phi}({\bar{\boldsymbol{e}}}^\top\boldsymbol{\phi} - V_{old}  )  + {\boldsymbol{e}}(\delta + V-V_{old})$
                  \STATE $
                        \bar{\boldsymbol{A}}  \leftarrow \bar{\boldsymbol{A}} \!- \alpha{\boldsymbol{\phi}}({\boldsymbol{\phi}}^\top \bar{\boldsymbol{A}}) $
                  \STATE $
                        {{\boldsymbol{\theta}}}  \leftarrow \bar{\boldsymbol{A}}{{\boldsymbol{\theta}}}+{\bar{\boldsymbol{e}}}$ 
                  \STATE $ V_{old} \leftarrow {V'}$
                  \STATE $\boldsymbol{\phi} \leftarrow {\boldsymbol{\phi}'}$
              \ENDWHILE
          \ENDFOR
     \end{algorithmic}
 \end{algorithm}
\section{True Online TD($\lambda$) as a Special Case of True Online TD($\lambda$)-Replan}
In this section we show that the true online TD($\lambda$) can be viewed as a special case of the true online TD($\lambda$)-Replan by fixing the weights used in the update rule (\ref{eq:Theta_incremental_update}).
\begin{theorem} \label{Theorem:2}
When $\theta_t = \theta_0$ the true online TD($\lambda$)-Replan algorithm reduces to the usual linear true online TD($\lambda$).
\end{theorem}
\begin{IEEEproof}
By defining $\bar{\boldsymbol{a}}_t:=\bar{\boldsymbol{A}}_t{\boldsymbol{\theta}}_0$, rule (\ref{eq:theta_incrmental}) becomes:
\begin{align}
\label{eq:fixed true online TD update}
    {{\boldsymbol{\theta}}}_{t+1} &= \bar{\boldsymbol{A}}_t{\boldsymbol{\theta}}_0+{\bar{\boldsymbol{e}}}_t \\
    {{\boldsymbol{\theta}}}_{t+1} &= \bar{\boldsymbol{a}}_t+{\bar{\boldsymbol{e}}}_t
\end{align}
In addition, $\bar{A}_t$ can be vectorized into $\bar{a}_t$ by multiplying (\ref{eq:Abar_t incremental}) by $\theta_0$ and substituting by the $\bar{\boldsymbol{a}}_t$ definition:
\begin{align}
\label{eq:abar_t incremental}
    \bar{\boldsymbol{a}}_t &= \bar{\boldsymbol{a}}_{t-1} - \alpha_t\boldsymbol{\phi}_t \left({\bar{\boldsymbol{a}}}_{t-1}^\top \boldsymbol{\phi}_t\right)
\end{align}
    Equation (\ref{eq:abar_t incremental}) can be combined with (\ref{eq:ebar_t incremental}) by simple addition of $\bar{\boldsymbol{a}}_t + \bar{\boldsymbol{e}}_t$ and substituting by ${\boldsymbol{\theta}}_{t+1}$ and substituting $\bar{\boldsymbol{a}}_{t-1} + \bar{\boldsymbol{e}}_{t-1}$ by ${\boldsymbol{\theta}}_{t}$ we have
\begin{align} \nonumber
   \bar{\boldsymbol{e}}_t + \bar{\boldsymbol{a}}_t &= {\bar{\boldsymbol{e}}}_{t-1} + \bar{\boldsymbol{a}}_{t-1} \\ \nonumber &- \alpha_t\boldsymbol{\phi}_t \left({\bar{\boldsymbol{a}}}_{t-1}^\top \boldsymbol{\phi}_t\right) + {\boldsymbol{e}}_{t}{\left(\delta_t + {{\boldsymbol{\theta}}}_{t}^\top{\boldsymbol{\phi}}_{t} - {{\boldsymbol{\theta}}}_{t-1}^\top{\boldsymbol{\phi}}_{t} \right)} 
        \\ \nonumber     &
        - \alpha_t\boldsymbol{\phi}_t \left(  {\bar{\boldsymbol{e}}}_{t-1}^\top\boldsymbol{\phi}_t - {{\boldsymbol{\theta}}}_{t-1}^\top{\boldsymbol{\phi}}_{t} \right)\\ \nonumber
   {\boldsymbol{\theta}}_{t+1} &= {{\boldsymbol{\theta}}}_{t} +  {\boldsymbol{e}}_{t}{\left(\delta_t + {{\boldsymbol{\theta}}}_{t}^\top{\boldsymbol{\phi}}_{t} - {{\boldsymbol{\theta}}}_{t-1}^\top{\boldsymbol{\phi}}_{t} \right)} \\
   \label{eq:true online TD update}
   & \qquad\qquad\qquad\qquad - \alpha_t\boldsymbol{\phi}_t \left( {{\boldsymbol{\theta}}}_{t}^\top{\boldsymbol{\phi}}_{t} -  {{\boldsymbol{\theta}}}_{t-1}^\top\boldsymbol{\phi}_t \right)
\end{align}
Update (\ref{eq:true online TD update}) along with (\ref{eq:e_t incrmental}) correspond exactly to the update rules of the true online TD($\lambda$) algorithm which conclude our proof.
\end{IEEEproof}
The TD($\lambda$)-Replan(1) replays all past experience in every step and as we said earlier it is on the far end of the spectrum of replaying. In order to reach a compromise that allows our algorithm to represents all range of replaying from non to full, we will look into reducing the complexity of our algorithm. So far, we have replayed all past experience by having current step replay rules to use past step weights as a base to update to current weights. However, our method allows for more flexibility. 

\section{True Online TD-Replan($\lambda$)}
The above methods constitutes the two ends of the spectrum of experience replay. The next normal development is to explore ways of representing all the spectrum between these two ends. In other words, we would like to construct a method that allows us to specify how much of a replay, if any, the agent should experience in each online step. This is an open question that one can address in several ways. One way to perform this requirement is by constructing a linear combination of the two methods using a hyper parameter $\acute{\lambda}$. Or, we can use the same $\lambda$  parametrisation in order to combine the depth of the target and the depth of the replay in one hyper parameter. To achieve this, we can alter the TD($\lambda$)-Replan method:
\begin{align} \nonumber
    {{\boldsymbol{\theta}}}_{t+1} &= \bar{\boldsymbol{A}}_t\left(\acute{\lambda}{\boldsymbol{\theta}}_t + \left(1-\acute{\lambda}\right){\boldsymbol{\theta}}_0 \right ) + {\bar{\boldsymbol{e}}}_t\\ \nonumber
    &= \bar{\boldsymbol{A}}_t\left(\acute{\lambda}{\boldsymbol{\theta}}_t + \left(1-\acute{\lambda}\right){\boldsymbol{\theta}}_0 \right ) + \left(\acute{\lambda}{\bar{\boldsymbol{e}}}_t + \left(1-\acute{\lambda}\right){\bar{\boldsymbol{e}}}_t \right )\\
    \label{eq:linear comb of Relan and TD}
    &= \acute{\lambda} (\bar{\boldsymbol{A}}_t{\boldsymbol{\theta}}_t + {\bar{\boldsymbol{e}}}_t) + (1-\acute{\lambda})(\bar{\boldsymbol{A}}_t{\boldsymbol{\theta}}_0 + {\bar{\boldsymbol{e}}}_t)
\end{align}
Since the last update is written as a linear combination of updates (\ref{eq:Theta_incremental_update})  and (\ref{eq:fixed true online TD update}) the resultant method is a parametrized combination of the two methods: the true online TD($\lambda$)-Replan and the true online TD($\lambda$). Therefor, the final method that we call TD($\lambda$)-Replan($\acute{\lambda}$) is defined by equations(\ref{eq:defAk}), (\ref{eq:linear comb of Relan and TD}), (\ref{eq:z_bar}), (\ref{eq:e}) and (\ref{eq:A_bar}).
Below we show a policy evaluation algorithm that is based on true online TD($\lambda$)-Replan($\acute{\lambda}$) method. 
\begin{algorithm}[H]
     \caption{true online TD($\lambda$)-Replan($\acute{\lambda}$) Learning} \label{algo:SarsaReplay(lambda)}
     \begin{algorithmic}
         \renewcommand{\algorithmicrequire}{\textbf{Input:}}
         \renewcommand{\algorithmicensure}{\textbf{Output:}}
         \REQUIRE $\alpha,\gamma,\lambda, \acute{\lambda}, {\boldsymbol{\theta}}_{init}$ \COMMENT {$\lambda$ may $= \acute{\lambda}$}
         \ENSURE  ${\boldsymbol{\theta}}$
          \STATE ${\boldsymbol{\theta}} \leftarrow {\boldsymbol{\theta}_{init}}$
          \FOR {all episodes}
              \STATE obtain initial state $\phi$
              \STATE ${\boldsymbol{\theta}}_{0} \leftarrow {\boldsymbol{\theta}}, {\boldsymbol{e}} \leftarrow \boldsymbol{0}, \bar{\boldsymbol{e}} \leftarrow 0, \bar{\boldsymbol{A}}  \leftarrow \mathcal{I}, V_{old}\leftarrow 0$
              \WHILE{$S$ is not Terminal}
                  \STATE obtain next feature vector $\acute{\phi}$                  
                 \STATE $ V \leftarrow {\boldsymbol{\theta}}^\top\boldsymbol{\phi}$
                  \STATE $ \acute{V} \leftarrow {\boldsymbol{\theta}}^\top\boldsymbol{\acute{\phi}}$
                  \STATE $ \delta \leftarrow R + \gamma \acute{V} -V$
                  \STATE $
                      {\boldsymbol{e}} \leftarrow {\boldsymbol{e}}\gamma\lambda -\alpha\boldsymbol{\phi}(\gamma\lambda{\boldsymbol{e}}^\top\boldsymbol{\phi} -1)$ 
                          \STATE $
                                {\bar{\boldsymbol{e}}} \leftarrow {\bar{\boldsymbol{e}}}  -\alpha\boldsymbol{\phi}({\bar{\boldsymbol{e}}}^\top\boldsymbol{\phi} - V_{old}  )  + {\boldsymbol{e}}(\delta + V-V_{old})$
                          \STATE $
                                \bar{\boldsymbol{A}}  \leftarrow \bar{\boldsymbol{A}} - \alpha{\boldsymbol{\phi}}({\boldsymbol{\phi}}^\top \bar{\boldsymbol{A}}) $
                          \STATE $
                                {{\boldsymbol{\theta}}}  \leftarrow \bar{\boldsymbol{A}}\left(\acute{\lambda}{{\boldsymbol{\theta}}} + (1-\acute{\lambda}){{\boldsymbol{\theta}}}_0\right)+{\bar{\boldsymbol{e}}}$ 
                  \STATE $ V_{old} \leftarrow \acute{V}$
                  \STATE $\boldsymbol{\phi} \leftarrow \acute{\boldsymbol{\phi}}$
              \ENDWHILE
          \ENDFOR
     \end{algorithmic}
 \end{algorithm}
 It should be noted that when true online TD($\lambda$)-Replan($\acute{\lambda}=0$) is exactly equivalent to the true online TD($\lambda$). It should be noted also that when we have $\acute{\lambda} = \lambda$ then the method is simply called TD-Replan($\lambda$).  Similar to true online TD($\lambda$), TD-Replan($0$) method is equivalent to TD($0$).
 \section{TD-REPLAN APPLIED ON RANDOM WALK}
\begin{figure}
  \centering
  {\includegraphics[width=4in,height=3in,clip,keepaspectratio]{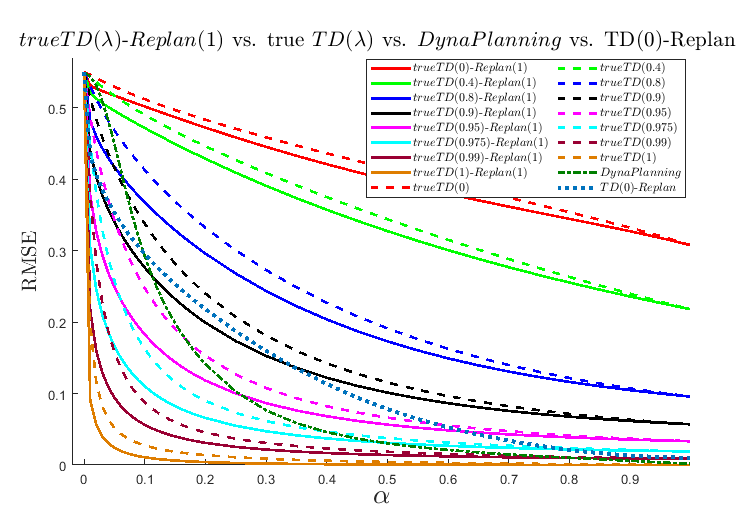}}
  \caption{Comparison of true online TD($\lambda$)-Replan(1) with true online TD($\lambda$), as well as TD(0)-Replan and Dyna Planning, on Random Walk on 17 of the first 10 episodes averaged over 20 trials for binary features. This shows the clear edge that our new method have over other methods despite the simplicity of the problem.}  \label{fig:TD-ReplanTrueTDReplayRandomWalkDisect}
\end{figure}

In this section we show the prediction performance of our algorithms on a random walk task for benchmarking. Random walk isolates the effect of the dynamic of the environment since selecting the actions is randomised based on a probability distribution that represents a fixed policy. This allows us to concentrate on the prediction capability of an algorithm. Our environment consists of 17 states \cite{SuttonBarto} and the process starts always form the far left hand side state and the episodes ends when the process reaches the far right state. In each step the action that moves the current state towards the right final state are given a fixed reward of $1/n$ where $n$ is the number of non terminal states. While, the action that moves the current state away from the right final state are given a reward of $-1/n$, the action that takes the process to the right final state is given a 0 reward, staying in the far left state is also given a reward of 0. Both actions have the same probability and no discount is used, i.e. $\gamma=1$. These setting allowed the RMSE error to be bounded to 1, and further allowed us to benchmark with other random walk problems. It can be easily proven that the sum of the rewards of each state can be analytically calculated to be $V(S_i)=(i-1)/n$.The features used are simple basis binary features that represents each state as a vector of zeros with one feature on at a time. 
Our experiments shows that our new methods TD-Replan have the least sensitivity to the step size and almost always guarantees convergence with maximum speed (in terms of number steps needed to converge).  Fig. \ref{fig:TD-ReplanTrueTDReplayRandomWalkDisect} shows that our algorithm outperforms the true online TD($\lambda$) for all $\lambda$ values in this domain. It also outperformed the TD(0)-Replan(1) algorithm \cite{TD-Replan(0)} as well as the Dyna Planning algorithm \cite{Dyna} both of which have a similar quadratic complexity, which shows that our algorithm clearly outperform those planning algorithms as well.

\begin{figure}
  \centering
  {\includegraphics[width=3.5in,height=2in,clip,keepaspectratio]{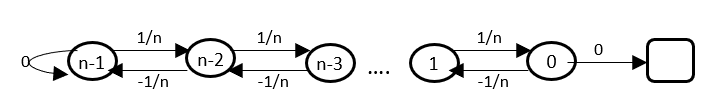}}
  \caption{Random Walk Task}  \label{fig:RandomWalkDisect}
\end{figure}

\section{DEEP SPARSE VARIATIONAL AUTOENCODER AND TD-Replan APPLIED ON MYOELECTRIC CONTROL}

In \cite{Myoelectric1} authors have shown how to control a two-dimension cursor via a set of surface electromyographic (sEMG) signals obtained from forearm activities. Fourteen abled-body subjects were studied one of which has a congenital upper-limb deficiency. They have conducted a set of myoelectric control experiments; their aim was to study the effect of arm position and donning/doffing of a textile hose that they used to obtain a set of sEMG signal readings. In their experiments, each subject controlled the cursor using a set of sixteen sEMG sensory signals attached to the subject's forearm. The task is to move a cursor from a location inside a circle that is presented to the subject on a computer screen to the centre of the circle using the sEMG signals coming from their muscles activities. In each experiment a set of sixteen pre-specified locations were randomly selected one after the other and the subject has to move the cursor to the requested location. 
The tasks were performed in approximation and sometimes the subjects failed to reach the target position in the allocated time. The subject is considered to hit the target if he/she sustained the cursor in a radius of 0.15 for 1 second. After the subject hits or misses the target, a pause of 1 second is enforced and the cursor is returned to the centre until all 16 targets have been presented (in random order). The dataset is publicly available in \cite{Myoelectric2}.


\begin{figure}
  \centering
  {\includegraphics[width=4in,height=2.5in,clip,keepaspectratio]{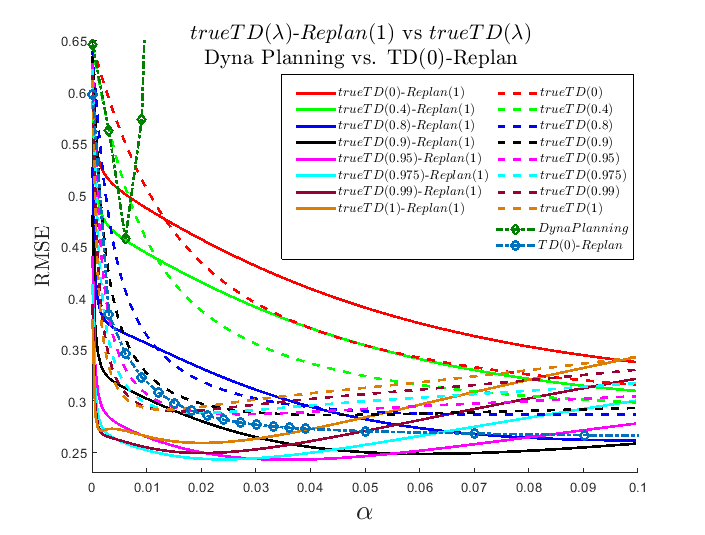}}
  \caption{Comparison of the RMSE for true online TD($\lambda$)-Replan(1) and true online TD($\lambda$) applied on Myoelectric control of cursor on a screen using normalised sEMG as features, all taken for the first 10 episodes and averaged over 66 trials with $\alpha$ values that spans 0.001 to 0.1 with 0.005 steps. the figure clearly shows that for high $\lambda$ values TD($\lambda$)-Replan(1) is more advantageous than true TD($\lambda$). We note that our algorithm has a wider maximal area and converges quicker to an optimal performance for a wider range of learning steps $\alpha$, making it more reliable and stable. Note that Dyna Planning has struggled to learn the environment's dynamics due to deep learning mapping the sEMG into a more elaborate but sparse space. On the other hand, TD(0)-Replan has performed relatively good as expected but could not outperform true TD(0.9)-Replan onward.}  \label{fig:Comparison on myoelectric no sAE}
\end{figure}

Our aim in this study is to show how to directly predict the future position of the cursor based on  current  raw sEMG arm signals, hence simplifying and transforming the ways in which such control models are constructed and trained. We will use our algorithms to predict the next position of the cursor on a screen based on the sEMG signals. The sEMG signal is packed with noise and variation that makes the task challenging. It has been attempted to be tackled using deep learning approach as in \cite{Li2018DeepRL} but in a supervised learning settings. In addition, we conduct a comparison study to show that our algorithm prediction accuracy can considerably outperform other widely used RL planning and non-planning algorithms such as the Dyna planning and true online TD($\lambda$) as well as TD(0)-Replan. All algorithms used the same two sets of features to compare their capacity. The first is the sEMG readings after normalisation, and the second is a set of features that are extracted from an auto encoder (AE). 

In \cite{Nexting1} authors showed that TD can predict the next sensor reading based on previous reading, they call it ‘nexting’. They have shown that nexting can be performed on a large number of sensory inputs to predict their next values in parallel. Their task was a  robot circling a pen continuously and their sensors (lights and ultrasonic) were predicted. In \cite{trueTD} authors have demonstrated how to predict two degrees of freedom task that involved the grip force and motor angle signals of a robotics hand. In this context, the sensory input plays the role of a reward. The point of view that rewards can be used to perform general prediction has been explored in several settings. For example, \cite{Nexting2} used a variety of stimuli as a reward function to learn animal behaviour and to model conditioning. 

\subsection{Deep Auto Encoder Structure and PreTraining}
The structure of the Sparse Auto Encoder is as follows. The encoder has 5 layers, the first treats the sEMG input as an image. The second, is a convolutional neural (CNN) layer that has 32 filters each of size 3x1 with a stride of 2 (yielding 8x32 features). This is followed by a relu which is then followed by another CNN layer that has a 64 filters each of size 3x1 with a stride of 2 (yielding an output of 4x64=256). This is followed by a relu layer which then is flattened into 256 neurons, which is followed by a fully connected layer to the 256 latent variables. No padding has been applied. The decoder mirrors these layers in the usual reversed manner (by using a transposed convolutional layers instead of the convolutional), both deconvolutional CNN layers have filter sizes of 2x1, no cropping has been applied. The sEMG signal has been treated as if it is a 2d gray scale image, the mission of the Sparse AE is to come up with a cleaner and a sparse decompressed representation of the sEMG signal, i.e. to map the 16 sEMG readings into $256=16^2$ features each specialised in a range of sEMG values.  
We start by training the AE in the usual unsupervised training fashion to learn the best representation for the sEMG sensor readings. We used a minibatch size of 512, the transfer function for all the encoders and decoders is the logistic sigmoid and the learning rate is set to $10^{-3}$. We have trained our deep learning feature extractor using all the episodes’ data regardless of the positions and trials (so all sixteen positions are considered). The input was normalised by re-scaling for each component of the 16 sEMG readings.  
After training the AE, we use the encoder to encode all sEMG signals in a predefined number of features that corresponds to the number of latent neurons.
The number of features used (without AE) is 16 and the model used one bias, while the number of latent features considered with AE is $16^2$. The number of episodes for training the AE was set to 10 where the loss was around 0.05. 
To show the prediction capabilities of our algorithm we use a value-function approach and we do not utilise direct parametrized policy search methods and our approach is a model-free RL. When we train our algorithm we do not try to build a model for the environment dynamics since it is not necessary to predict the value function (or to do a policy search, both of these can be done without the environment model by sampling interaction between the agent and its environment). However, we will compare the performance of our TD-Replan with Dyna Planning that create a model that predicts the next state and next reward based on current state \cite{Dyna}.

\subsection{True Online TD-Replan Training Results}
The data set has been divided into trials each trial consists 10 random episodes that belongs to the same task (starting position for the cursor), each episodes constitutes a task of moving the cursor from a start position to the centre of a circle on the screen using the subject sEMG signal. So the number of steps of each episodes varies. In order to train a network to perform a prediction for a task, we have bundled the episodes related to each task together.
The model has been left to run to the end of each episode without a stop condition to capture the full experience. The reward signal is taken to be the normalised difference between current position and the target position for the X coordinate and the Y coordinate separately. The results are shown for predicting the X signal for brevity. All episodes of the six starting positions that have significant variation along the X axes are considered.
We have not used the calibration runs of the subjects (the first 9) since during calibration one of the coordinated were artificially set to 0 and the other is set to a fixed number and the cursor was not actually moved using the sEMG \cite{Myoelectric3}. Runs 22 onward were not included due to the deterioration of performance of the participants due to donning/doffing effect, therefore all runs 10-21 were utilized. All experiments were done on an online fashion and all arms positions were considered equally without distinction. The maximum number of trials is 66 (11 for each task) and all of them has been utilised to obtain the averages (in our settings a trail is a set of 10 episodes).  $\gamma$ was set to 0.95 and we have used the same representation across the tasks. 

Fig. \ref{fig:Comparison on myoelectric no sAE} shows the comparison for the normalised sEMG features, where no deep learning feature extractor is employed. This figure shows that our algorithm performance exceeds the performance of the true online TD for any relatively high $\lambda$ values ($\ge 0.8$) specifically at high $\alpha$ values $\ge 0.05$. 

\begin{figure}
  \centering
    {\includegraphics[width=3.8in,height=3.2in,clip,keepaspectratio]{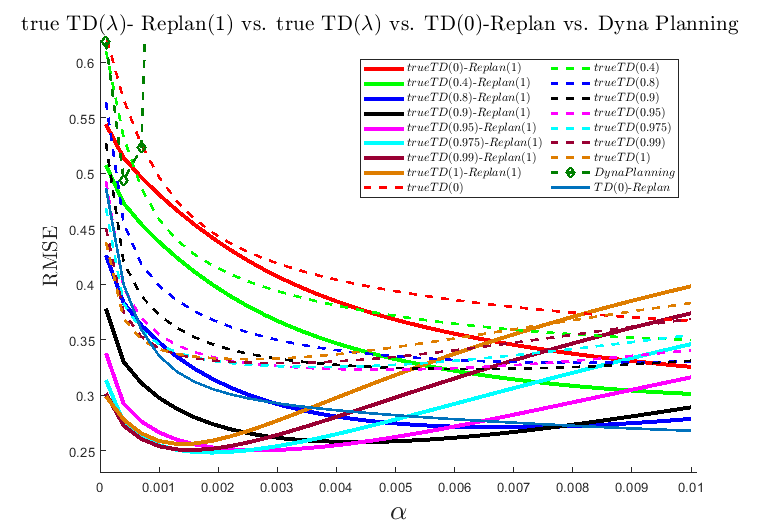}}
    \caption{RMSE comparison for true online TD($\lambda$)-Replan(1), true online TD($\lambda$), TD(0)-Replan and Dyna Planning. The methods are applied on myoelectric control of cursor on a screen, where 16 normalised sEMG are fed into a Sparse Auto Encoder to extract a more elaborate set of features ($16^2$). All results are taken for the first 10 episodes and averaged over 66 trials with $\alpha$ values that spans $10^{-4}$ to $10^{-3}$ with $3\times10^{-4}$  increment. The figure clearly shows that when using deeply learned features TD($\lambda$)-Replan(1) outperforms the true TD($\lambda$) for all $\lambda$ values with a considerable margin.}
     \label{fig:sAE TD(lambda)-Replay(1)} 
\end{figure}

Fig. \ref{fig:sAE TD(lambda)-Replay(1)} 
shows clearly that our algorithm outperforms the true online TD($\lambda$) for all $\lambda$ values in this domain for the deep extracted features. The figure shows that the difference between our algorithm and the other algorithms becomes more prominent, demonstrating the suitability of our algorithm to this type of deep learning extraction. We note that true online TD($\lambda$)-Replan converges quicker to an optimal performance making it agile. Another important property to note, is that the algorithm starts almost readily with low RMSE levels when we increase the replay depth, and quickly converges to its optimal performance for small to intermediate learning step. This demonstrate that our algorithm suitable for real time and critical applications that needs minimal training and quick response. Note that $\lambda$ performed best for 0.9 as is normally expected.

\begin{figure}
        {\includegraphics[width=3.8in,height=3.2in,clip,keepaspectratio]{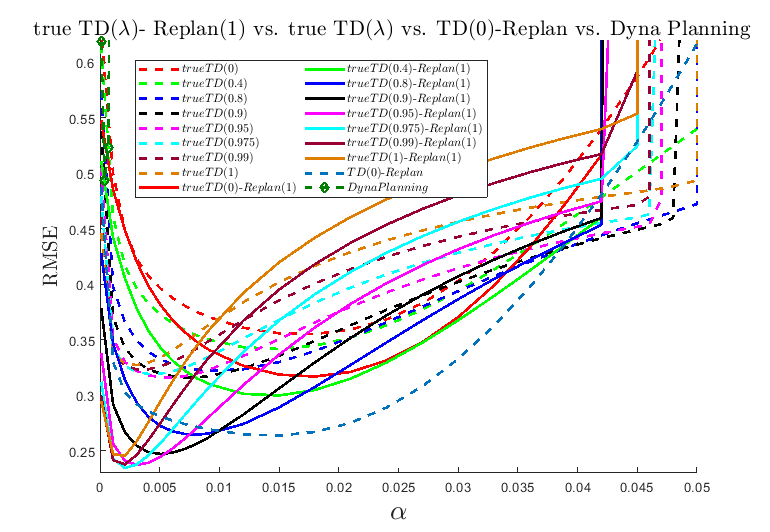}}
    \caption{Same as for Fig. \ref{fig:sAE TD(lambda)-Replay(1)} but over a wider range of values of $\alpha$. It shows that for this type of features the smaller $\alpha$ values generates better results for both algorithms.}  
    \label{fig:sAE TD(lambda)-Replay(1) wider range}       
\end{figure}

Fig. \ref{fig:sAE TD(lambda)-Replay(1) wider range} shows that the values that keep all methods convergent are the range shown in Fig. \ref{fig:sAE TD(lambda)-Replay(1)} over which TD($\lambda$)-Replan(1) outperformed all other methods. Note that Dyna Planning is included but hardly can be seen due to its divergence for $\alpha$ values beyond 0.002.

\begin{figure}
  \centering
  {\includegraphics[width=3.8in,height=3.8in,clip,keepaspectratio]{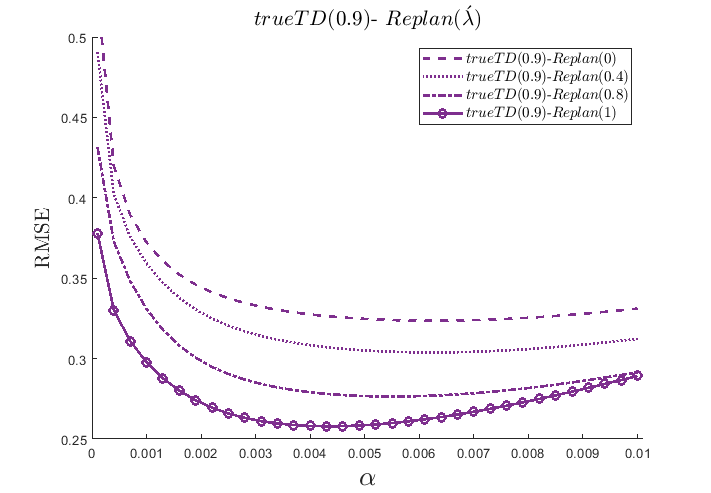}} 
  \caption{RMSE Comparison for true online TD($0.9$)-Replan($\acute{\lambda}$) applied on myoelectric control of cursor on a screen with a $16^2$ latent features coming from an encoder, all taken for the first 10 episodes and averaged over 66 trials with $\alpha$ values that spans $10^{-4}$ to $10^{-3}$. The figure clearly shows that increasing the replay depth $\acute{\lambda}$ increases the algorithm performance when using deeply learned features. Note that the same applies for other $\lambda$ values but is not shown for brevity.}  \label{fig:sAE TD(0.9)-Replay(lambda)}
\end{figure}

Fig. \ref{fig:sAE TD(0.9)-Replay(lambda)} shows the the comparison for true online TD($\lambda=0.9$)-Replan($\acute{\lambda}$) with different $\acute{\lambda}$, demonstrating that the depth of the replay plays an important role when the features are deep and the rewards is not delayed. 
We should note also that true online TD($\lambda$)-Replan(0) = true online TD($\lambda$) as per Theorem \ref{Theorem:2} and hence the latter is actually included in the comparison. We can see the theme of a stable optimal performance for a range of $\alpha$ values when we increase the replay depth (i.e. when we increase $\acute{\lambda}$) up to 0.8. For $\acute{\lambda}=1$ the algorithm increased its performance for the smaller $\alpha$ values but then started to decrease for relatively higher values. This type of behaviour is expected for this hyper parameter, resembling the usual $\lambda$ behaviour.

\section{Conclusion}
In this paper, we have introduced a novel reinforcement learning method, namely the true online TD($\lambda$)-Replan($\acute{\lambda}$) that extends the capabilities of the true online TD($\lambda$) method to allow the method to replay all or part of its past experience according to the $\acute{\lambda}$ parameter. In addition, $\lambda$ allows it to choose the depth of its targets as per the norm for TD($\lambda$) methods. The cost of the algorithm is quadratic in the number of features and the algorithm is suitable for planning. Further, we proved that the true online TD($\lambda$) method becomes a special case of our method. This in turn allowed us to design an algorithm that can scan the full spectrum between full and partial planning based on the suitability of the application and the response and complexity requirements. Specifically, true online TD($\lambda$) = true online TD($\lambda$)-Replan(0). We have tested the efficacy of our algorithm on two benchmarking domains, in one of which we have combined our algorithm with a sparse autoencoder that utilises CNN layers. Both domains confirmed the utility and high performance of our algorithm in comparison to other more expensive algorithms. Further, the results show that our algorithm constituted a good match for a deep learning extractor, paving the way for further integration in the future. Future work includes showing that our methods can be integrated with on-policy or off-policy learning updates to produce new control methods, in addition to tackling an end-to-end training of a deep reinforcement learning model that is based on our method. 

\ifCLASSOPTIONcaptionsoff
  \newpage
\fi

\bibliographystyle{IEEEtran}

\bibliography{trueTDReplay}

\end{document}